# Can We Learn to Beat the Best Stock

**Allan Borodin**                                    BOR@CS.TORONTO.EDU
*Department of Computer Science*
*University of Toronto*
*Toronto, ON, M5S 3G4 Canada*

**Ran El-Yaniv**                                    RANI@CS.TECHNION.AC.IL
*Department of Computer Science*
*Technion - Israel Institute of Technology*
*Haifa 32000, Israel*

**Vincent Gogan**                                    VINCENT@CS.TORONTO.EDU
*Department of Computer Science*
*University of Toronto*
*Toronto, ON, M5S 3G4 Canada*

## Abstract

A novel algorithm for actively trading stocks is presented. While traditional expert advice and "universal" algorithms (as well as standard technical trading heuristics) attempt to predict winners or trends, our approach relies on predictable statistical relations between all pairs of stocks in the market. Our empirical results on historical markets provide strong evidence that this type of technical trading can "beat the market" and moreover, can beat the best stock in the market. In doing so we utilize a new idea for smoothing critical parameters in the context of expert learning.

## 1. Introduction

The portfolio selection (PS) problem is a challenging problem for machine learning, online algorithms and, of course, computational finance. As is well known (e.g. see Lugosi, 2001) sequence prediction under the log loss measure can be viewed as a special case of portfolio selection, and perhaps more surprisingly, from a certain worst case minimax criterion, portfolio selection is not essentially any harder (than prediction) as shown in (Cover & Ordentlich, 1996) (see also Lugosi, 2001, Thm. 20 & 21). But there seems to be a qualitative difference between the practical utility of "universal" sequence prediction and "universal" portfolio selection. Simply stated, universal sequence prediction algorithms under various probabilistic and worst-case models appear to work very well in practice whereas the known universal portfolio selection algorithms do not seem to provide any substantial benefit over a naive investment strategy (see Section 5).

A major pragmatic question is whether or not a computer program can consistently outperform the market. A closer inspection of the interesting ideas developed in information theory and online learning suggests that a promising approach is to exploit the natural volatility in the market and in particular to benefit from simple and rather persistent statistical relations between stocks rather than to try to predict stock prices or "winners".

We present a non-universal portfolio selection algorithm[1], which does not try to predict winners. The motivation behind our algorithm is the rationale behind *constant rebalancing* algorithms and the worst case study of universal trading introduced by Cover (1991). Not only does our proposed algorithm substantially "beat the market" on historical markets, it also beats the best stock. So why are we presenting this algorithm and not just simply making money? There are, of course some caveats and obstacles to utilizing the algorithm. But for large investors the possibility of a goose laying silver (if not golden) eggs is not perhaps impossible.

## 2. The Portfolio Selection Problem

Assume a market with $m$ stocks. Let $\mathbf{v}_t = (v_t(1), \ldots, v_t(m))$ be the daily closing prices[2] of the $m$ stocks for the $t^{th}$ day, where $v_t(j)$ is the price of the $j$th stock. It is convenient to work with *relative prices* $x_t(j) = v_t(j)/v_{t-1}(j)$ so that an investment of \$$d$ in the $j$th stock just before the $t^{th}$ day yields $dx_t(j)$ dollars. We let $\mathbf{x}_t = (x_t(1), \ldots, x_t(m))$ denote the *market vector* of relative prices corresponding to the $t^{th}$ day. A *portfolio* $\mathbf{b}$ is an allocation of wealth in the stocks, specified by the proportions $\mathbf{b} = (b(1), \ldots, b(m))$ of current dollar wealth invested in each of the stocks, where $b(j) \geq 0$ and $\sum_j b(j) = 1$. The *daily return* of a portfolio $\mathbf{b}$ w.r.t. a market vector $\mathbf{x}$ is $\mathbf{b} \cdot \mathbf{x} = \sum_j b(j)x(j)$ and the (compound) *total return*, $\mathsf{ret}_X(\mathbf{b}_1, \ldots, \mathbf{b}_n)$, of a sequence of portfolios $\mathbf{b}_1, \ldots, \mathbf{b}_n$ w.r.t. a market sequence $X = \mathbf{x}_1, \ldots, \mathbf{x}_n$ is $\prod_{t=1}^n \mathbf{b}_t \cdot \mathbf{x}_t$. A portfolio selection algorithm $\mathcal{A}$ is any deterministic or randomized rule for specifying a sequence of portfolios and we let $\mathsf{ret}_X(\mathcal{A})$ denote its total return for the market sequence $X$.

The simplest strategy is to "buy-and-hold" stocks using some portfolio $\mathbf{b}$. We denote this strategy by BAH$_\mathbf{b}$ and let U-BAH denote the uniform buy-and-hold when $\mathbf{b} = (1/m, \ldots, 1/m)$. We say that a portfolio selection algorithm "beats the market" when it outpeforms U-BAH on a given market sequence although in practice "the market" can be represented by some non-uniform BAH.[3] Buy-and-hold strategies rely on the tendency of successful markets to grow. Much of modern portfolio theory focuses on how to choose a good $\mathbf{b}$ for the buy-and-hold strategy. The seminal ideas of Markowitz (1959) yield an algorithmic procedure for choosing the weights of the portfolio $\mathbf{b}$ so as to minimize the variance for any feasible expected return. This variance minimization is possible by placing appropriate (larger) weights on subsets of sufficiently anti-correlated stocks, an idea which we shall also utilize. We denote the optimal *in hindsight* buy-and-hold strategy (i.e. invest only in the best stock) by BAH*.

An alternative approach to the static buy-and-hold is to dynamically change the portfolio during the trading period. This approach is often called "active trading". One example of active trading is *constant rebalancing*; namely, fix a portfolio $\mathbf{b}$ and (re)invest your dollars *each day* according to $\mathbf{b}$. We denote this constant rebalancing strategy by CBAL$_\mathbf{b}$ and let CBAL* denote the optimal (in hindsight) CBAL. A constant rebalancing strategy can often

---

1. Any PS algorithm can be modified to be universal by investing any fixed fraction of the initial wealth in a universal algorithm.

2. There is nothing special about "daily closing prices" and the problem can be defined with respect to any (sub)sequence of the (intra-day) sequence of all price offers which appear in the stock market.

3. For example the Dow Jones Industrial Average (DJIA) is calculated as a non uniform average of the 30 DJIA stocks; see e.g. `http://www.dowjones.com/`





take advantage of market fluctuations to achieve a return significantly greater than that of BAH*. CBAL* is always at least as good as the best stock BAH* and in some real market sequences a constant rebalancing strategy will take advantage of market fluctuations and significantly outperform the best stock (see e.g. Table 1). For now, consider Cover and Gluss's (1986) classic (but contrived) example of a market consisting of cash and one stock and the market sequence of price relatives $\binom{1}{1/2}, \binom{1}{2}, \binom{1}{1/2}, \binom{1}{2}, \ldots$. Now consider the CBAL$_{\mathbf{b}}$ with $\mathbf{b} = (\frac{1}{2}, \frac{1}{2})$. On each odd day the daily return of CBAL$_{\mathbf{b}}$ is $\frac{1}{2}1 + \frac{1}{2}\frac{1}{2} = \frac{3}{4}$ and on each even day, it is $3/2$. The total return over $n$ days is therefore $(9/8)^{n/2}$, illustrating how a constant rebalancing strategy can yield exponential returns in a "no-growth market". Under the assumption that the daily market vectors are observations of identically and independently distributed (i.i.d) random variables, it is shown in (Cover & Thomas, 1991) that CBAL* performs at least as good (in the sense of expected total return) as the best online portfolio selection algorithm. However, many studies (see e.g. Lo & MacKinlay, 1999) argue that stock price sequences do have long term memory and are not i.i.d.

A non-traditional objective (in computational finance) is to develop online trading strategies that are in some sense *always guaranteed* to perform well.[4] Within a line of research pioneered by Cover (Cover & Gluss, 1986; Cover, 1991; Cover & Ordentlich, 1996) one attempts to design portfolio selection algorithms that can provably do well (in terms of their total return) with respect to some online or offline benchmark algorithms. Two natural online benchmark algorithms are the uniform buy and hold U-BAH, and the uniform constant rebalancing strategy U-CBAL, which is CBAL$_{\mathbf{b}}$ with $\mathbf{b} = (\frac{1}{m}, \ldots, \frac{1}{m})$. A natural offline benchmark is BAH* and a more challenging offline benchmark is CBAL*.

A portfolio selection algorithm $\mathcal{A}$ is called *universal* if *for every market sequence X* over $n$ days, it guarantees a subexponential ratio (in $n$) between its return $\mathsf{ret}_X(\mathcal{A})$ and that of $\mathsf{ret}_X(\text{CBAL}^*)$. In particular, Cover and Ordentlich's *Universal Portfolios* algorithm (Cover, 1991; Cover & Ordentlich, 1996), denoted here by UNIVERSAL, was proven to be universal; more specifically *for every market sequence X* of $m$ stocks over $n$ days, it guarantees the subexponential (indeed polynomial) ratio

$$\mathsf{ret}_X(\text{CBAL}^*)/\mathsf{ret}_X(\text{UNIVERSAL}) = O\left(n^{\frac{m-1}{2}}\right). \tag{1}$$

From a theoretical perspective this is surprising as this performance ratio is bounded by a polynomial in $n$ (for fixed $m$) whereas CBAL* is capable of exponential returns. From a practical perspective, this bound is not very useful because the empirical returns observed for CBAL* portfolios is often not exponential in the number of trading days. However, the motivation that underlies the potential of CBAL algorithms is useful! We follow this motivation and develop a new algorithm which we call ANTICOR. By attempting to systematically follow the constant rebalancing philosophy, ANTICOR is capable of some extraordinary performance in the absence of transaction costs, or even with very small transaction costs.

---

4. A trading strategy is *online* if it computes the portfolio for the $(t+1)^{st}$ day using only market information for the first $t$ days. This is in contrast to *offline* algorithms such as U-BAH*, CBAL* and the optimal strategy of picking the best stock for each individual day. Such offline algorithms compute a sequence of portfolios as a function of the entire market sequence.





## 3. Trying to Learn the Winners

The most direct approach to expert learning and portfolio selection is a "(reward based) weighted average prediction" scheme, which adaptively computes a weighted average of experts by gradually increasing (by some multiplicative or additive update rule) the relative weights of the more successful experts. In this section we briefly discuss some related portfolio selection results along these lines.

For example, in the context of the PS problem consider the "exponentiated gradient" EG($\eta$) algorithm proposed by (Helmbold et al., 1998). The EG($\eta$) algorithm computes the next portfolio to be

$$\mathbf{b}_{t+1}(j) = \frac{\mathbf{b}_t(j) \exp \left\{ \eta \mathbf{x}_t(j)/(\mathbf{b}_t \cdot \mathbf{x}_t) \right\}}{\sum_{j=1}^m \mathbf{b}_t(j) \exp \left\{ \eta \mathbf{x}_t(j)/(\mathbf{b}_t \cdot \mathbf{x}_t) \right\}},$$

where $\eta$ is a "learning rate" parameter. EG was designed to greedily choose the best portfolio for yesterday's market $\mathbf{x}_t$ while at the same time paying a penalty from moving far from yesterday's portfolio. For a universal bound on EG, Helmbold *et al.* set $\eta = 2x_{\min} \sqrt{2(\log m)/n}$ where $x_{\min}$ is a lower bound on any price relative.[5] It is easy to see that as $n$ increases, $\eta$ decreases to 0 so that we can think of $\eta$ as being very small in order to achieve universality. When $\eta = 0$, the algorithm EG($\eta$) degenerates to the uniform CBAL (assuming we started with a uniform portfolio) which is not a universal algorithm. It is also the case that if each day the price relatives for all stocks were identical, then EG (as well as other PS algorithms) will converge to the uniform CBAL. Combining a small learning rate with a "reasonably balanced" market we expect the performance of EG to be similar to that of the uniform CBAL and this is confirmed by our experiments (see Table 1).[6]

Cover's universal algorithms adaptively learn each day's portfolio by increasing the weights of successful CBALs. The update rule for these universal algorithms is

$$\mathbf{b}_{t+1} = \frac{\int \mathbf{b} \cdot \mathsf{ret}_t(\mathrm{CBAL}_{\mathbf{b}}) d\mu(\mathbf{b})}{\int \mathsf{ret}_t(\mathrm{CBAL}_{\mathbf{b}}) d\mu(\mathbf{b})},$$

where $\mu(\cdot)$ is some prior distribution over portfolios. Thus, the weight of a possible portfolio is proportional to its total return $\mathsf{ret}_t(\mathbf{b})$ thus far times its prior. The particular universal algorithm we consider in our experiments uses the Dirichlet prior (with parameters $(\frac{1}{2}, \ldots, \frac{1}{2})$) (Cover & Ordentlich, 1996).[7] Somewhat surprisingly, as noted in (Cover & Ordentlich, 1996) the algorithm is equivalent to a static weighted average (given by $\mu(\mathbf{b})$) over all CBALs (see also Borodin & El-Yaniv, 1998, p. 291). This equivalence helps to demystify the universality result and also shows that the algorithm can never outperform CBAL*.

---

5. Helmbold *et al.* show how to eliminate the need to know $x_{\min}$ and $n$. While EG can be made universal, its performance ratio is only sub-exponential (and not polynomial) in $n$.

6. Following Helmbold *et al.* we fix $\eta = 0.01$ in our experiments. Additional experiments, for a wide range of fixed $\eta$ settings, confirm that for our datasets the choice of $\eta = 0.01$ is an optimal or near optimal choice. Of course, it is possible to adaptively set $\eta$ throughout the trading period, but that is beyond the scope of this paper.

7. The papers (Cover, 1991; Cover & Ordentlich, 1996; Blum & Kalai, 1998) consider a simpler version of this algorithm where the (Dirichlet) prior is uniform. This algorithm is also universal and achieves a ratio $\Theta(n^{m-1})$. Experimentally (on our datasets) there is a negligible difference between these two variants and here we only report on the results of the asymptotically optimal algorithm.





A different type of "winner learning" algorithm can be obtained from any sequence prediction strategy, as noted in (Borodin, El-Yaniv, & Gogan, 2000). For each stock $j$, a (soft) sequence prediction algorithm provides a probability $p(j)$ that the next symbol will be $j \in \{1, \ldots, m\}$. We view this as a prediction that stock $j$ will have the best relative price for the next day and set $\mathbf{b}_{t+1}(j) = p_j$. The paper (Borodin et al., 2000) considers predictions made using the prediction component of the well-known Lempel-Ziv (LZ) lossless compression algorithm (Ziv & Lempel, 1978). This prediction component is nicely described in (Langdon, 1983) and in (Feder, 1991). As a prediction algorithm, LZ is provably powerful in various senses. First it can be shown that it is asymptotically optimal with respect to any stationary and ergodic finite order Markov source (Rissanen, 1983; Ziv & Lempel, 1978). Moreover, Feder shows that LZ is also universal in a worst case sense with respect to the (offline) benchmark class of all finite state prediction machines. To summarize, the common approach to devising PS algorithms has been to attempt and learn winners using simple or more sophisticated winner learning schemes.

## 4. The Anticor Algorithm

We propose a different approach, motivated by a CBAL-inspired "philosophy". How can we interpret the success of the uniform CBAL on the Cover and Gluss example of Section 2? Clearly, the uniform CBAL here is taking advantage of price fluctuation by constantly transferring wealth from the high performing stock to the relatively low performing stock. Even in a less contrived market, a CBAL is capable of large returns. A market model favoring the use of a CBAL is one in which stock growth rates are stable in the long term and occasional larger return rates will be followed by smaller rates (and vice versa). This market phenomenon is is sometimes called "reversal to the mean".

There are many ways that one can interpret and implement algorithms based on the philosophy of "reversal to the mean". In particular, any CBAL can be viewed as a static implementation of this philosophy. We now describe the motivation and basic ingredients in our ANTICOR algorithm which adaptively (based on recent empirical statistics) and rather aggressively[8] implements "reversal to the mean".

For a given trading day, consider the most recent past $w$ trading days, where $w$ is some integer parameter. The growth rate of any stock $i$ during this *window* of time is measured by the product of relative prices during this window.[9] Motivated by the assumption that we have a portfolio of stocks that are all performing similarly in terms of long term growth rates, ANTICOR's first condition for transferring money from stock $i$ to stock $j$ is that the growth rate for stock $i$ exceeds that of stock $j$ in this most recent window of time.[10] In addition, the ANTICOR algorithm requires some indication that stock $j$ will start to emulate the past growth of stock $i$ in the near future. To this end, ANTICOR requires a positive correlation between stock $i$ during the second last window and stock $j$ during the last window. The relative extent to which we will transfer money from stock $i$ to stock $j$ will depend on

---

8. Our ANTICOR algorithm is aggressive (say, compared to CBAL) in the sense that it can transfer all assets out of a given stock. Various heuristics can be used to moderate this behavior.

9. Since we would rather deal with arithmetic instead of geometric means we will use the logarithms of relative prices.

10. Note that here the underlying model assumption is reversal to the *same* mean. One can modify the algorithm so as to account for different means.





the strength of this correlation as well as the strength of the "self anti-correlations" for stocks $i$ and $j$ (again in two consecutive windows). ANTICOR is so named because we use these correlations and anticorrelations in consecutive windows to indicate the potential for anticorrelations of the growth rates for stocks $i$ and $j$ in the near future (with hopefully the growth rate of stock $j$ becoming greater than that of stock $i$).

Formally, we define

$$\mathsf{LX}_1 = \log(\mathbf{x}_{t-2w+1})^T, \ldots, \log(\mathbf{x}_{t-w})^T \quad \text{and} \quad \mathsf{LX}_2 = \log(\mathbf{x}_{t-w+1})^T, \ldots, \log(\mathbf{x}_t)^T, \tag{2}$$

where $\log(\mathbf{x}_k)$ denotes $(\log(x_k(1)), \ldots, \log(x_k(m)))$. Thus, $\mathsf{LX}_1$ and $\mathsf{LX}_2$ are the two vector sequences (equivalently, two $w \times m$ matrices) constructed by taking the logarithm over the market subsequences corresponding to the time windows $[t-2w+1, t-w]$ and $[t-w+1, t]$, respectively. We denote the $j$th column of $\mathsf{LX}_k$ by $\mathsf{LX}_k(j)$. Let $\mu_k = (\mu_k(1), \ldots, \mu_k(m))$, be the vectors of averages of columns of $\mathsf{LX}_k$. Similarly, let $\sigma_k$, be the vector of standard deviations of columns of $\mathsf{LX}_k$. The cross-correlation matrix (and its normalization) between column vectors in $\mathsf{LX}_1$ and $\mathsf{LX}_2$ are defined as[11]

$$M_{cov}(i,j) = \frac{1}{w-1}(\mathsf{LX}_1(i) - \mu_1(i))^T(\mathsf{LX}_2(j) - \mu_2(j));$$
$$M_{cor}(i,j) = \begin{cases} \frac{M_{cov}(i,j)}{\sigma_1(i)\sigma_2(j)} & \sigma_1(i), \sigma_2(j) \neq 0; \\ 0 & \text{otherwise.} \end{cases} \tag{3}$$

$M_{cor}(i,j) \in [-1, 1]$ measures the correlation between log-relative prices of stock $i$ over the first window and stock $j$ over the second window. We note that if $\sigma_1(i)$ (respectively, $\sigma_2(j)$) is zero over some window then the growth rate of stock $i$ during the second last window (respectively, stock $j$ during the last window) is constant during this window. For sufficiently large windows of time constant growth of any stock $i$ is unlikely. However, in this unlikely case we choose not to move money into or out of such a stock $i$.[12]

For each pair of stocks $i$ and $j$ we compute $\mathsf{claim}_{i \to j}$, the extent to which we want to shift our investment from stock $i$ to stock $j$. Namely, there is such a claim iff $\mu_2(i) > \mu_2(j)$ and $M_{cor}(i,j) > 0$ in which case $\mathsf{claim}_{i \to j} = M_{cor}(i,j) + A(i) + A(j)$ where $A(h) = |M_{cor}(h,h)|$ if $M_{cor}(h,h) < 0$, else 0. Following our interpretation for the success of a CBAL, $M_{cor}(i,j) > 0$ is used to predict that stocks $i$ and $j$ will be correlated in consecutive windows (i.e. the current window and the next window based on the evidence for the last two windows) and $M_{cor}(h,h) < 0$ predicts that stock $h$ will be negatively auto-correlated over consecutive windows. Finally, $\mathbf{b}_{t+1}(i) = \mathbf{b}_t(i) + \sum_{j \neq i}[\mathsf{transfer}_{j \to i} - \mathsf{transfer}_{i \to j}]$ where $\mathsf{transfer}_{i \to j} = \mathbf{b}_t(i) \cdot \mathsf{claim}_{i \to j} / \sum_j \mathsf{claim}_{i \to j}$. A pseudocode summarizing the ANTICOR algorithm appears in Figure 1. The pseudocode describes the routine ANTICOR$(w, t, X_t, \hat{\mathbf{b}}_t)$ that receives a window size $w$, the current trading day $t$, the historical market sequence $X_t$ (giving the market vectors corresponding to days $1, \ldots, t$) and the current portfolio $\hat{\mathbf{b}}_t$ defined to be $\hat{\mathbf{b}}_t = \frac{1}{\mathbf{b}_t \cdot \mathbf{x}_t}(\mathbf{b}_t(1)\mathbf{x}_t(1), \ldots, \mathbf{b}_t(m)\mathbf{x}_t(m))$. The routine is first called with an empty historical market sequence and with $\hat{\mathbf{b}}_t$ being the uniform portfolio (over $m$ stocks). The routine returns the new portfolio, to which we should rebalance at the start of the $(t+1)^{st}$ trading day.

---

11. Recall that the *correlation coefficient* is a normalized covariance with the covariance divided by the product of the standard deviations; that is, $Cor(X, Y) = Cov(X, Y)/(std(X) * std(Y))$ where $Cov(X, Y) = E[(X - mean(X))(Y - mean(Y))]$.

12. Of course, other approaches can be used to accommodate constant or nearly constant growth rate.

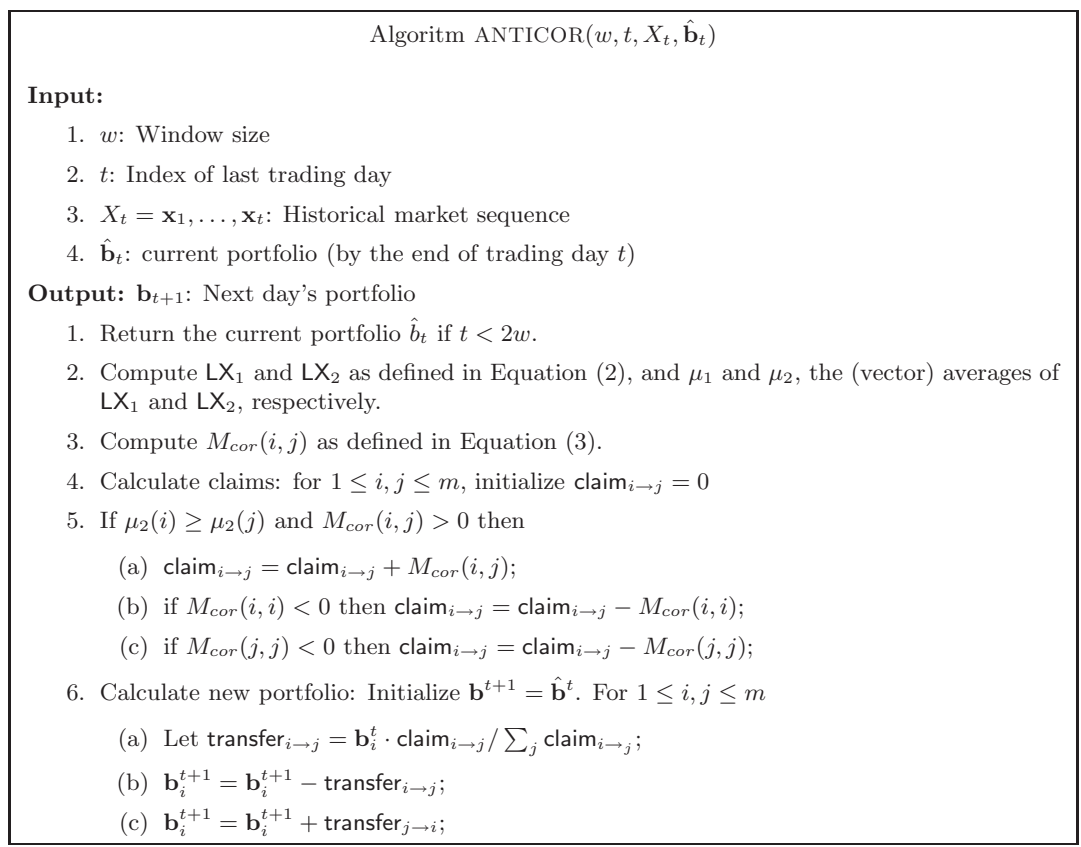

Figure 1: Algorithm ANTICOR

Our ANTICOR$_w$ algorithm has one critical parameter, the window size $w$. In Figure 2 we depict the total return of ANTICOR$_w$ on two historical datasets as a function of the window size $w = 2, \ldots, 30$ (detailed descriptions of these datasets appear in Section 5). As we might expect, the performance of ANTICOR$_w$ depends significantly on the window size. However, for all $w$, ANTICOR$_w$ beats the uniform market and, moreover, it beats the best stock using most window sizes. Of course, in online trading we cannot choose $w$ in hindsight. Viewing the ANTICOR$_w$ algorithms as experts, we can try to learn the best expert. But the windows, like individual stocks, induce a rather volatile set of experts and standard expert combination algorithms (Cesa-Bianchi et al., 1997) tend to fail.[13]

Alternatively, we can adaptively learn and invest in some weighted average of all ANTICOR$_w$ algorithms with $w$ less than some maximum $W$. The simplest case is a uniform investment on all the windows; that is, a uniform buy-and-hold investment on the algorithms ANTICOR$_w$, $w \in [2, W]$, denoted by BAH$_W$(ANTICOR). Figure 3 graphs the total return of BAH$_W$(ANTICOR) as a function of $W$ for all values of $2 \le W \le 50$ for the four datasets we consider here. Considering these graphs, our choice of $W = 30$ was arbitrary but clearly not optimal. Of course, we could try to optimize the parameter $W$ for each particular dataset by training the algorithm on historical data before beginning to trade. However, our claim is that almost any choice of $W$ will yield returns that beat the best stock (the only exception is $W = 2$ in the DJIA dataset).

---

13. This assertion is based on empirical studies we conducted with various 'expert advice' algorithms.

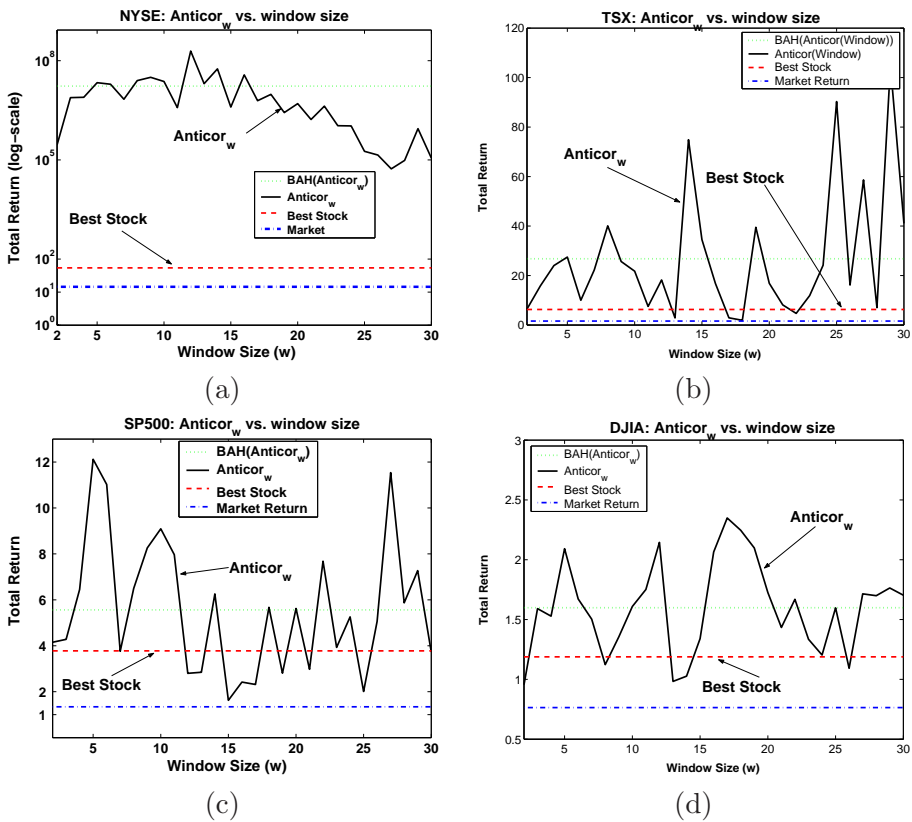

Figure 2: ANTICOR$_w$'s total return (per \$1 investment) vs. window size $2 \leq w \leq 30$ for (a) NYSE; (b) TSX; (c) SP500; (d) DJIA. The dashed (red) lines represent the final return of the best stock and the dash-dotted (blue) lines, the final return the (uniform) market. The dotted (green) horizontal lines represent a uniform investment on a number of ANTICOR$_w$ applications as later described.

Since we now consider the various algorithms as stocks (whose prices are determined by the cumulative returns of the algorithms), we are back to our original portfolio selection problem and if the ANTICOR algorithm performs well on stocks it may also perform well on algorithms. We thus consider active investment in the various ANTICOR$_w$ algorithms using ANTICOR. We again consider all windows $w \leq W$. Of course, we can continue to compound the algorithm any number of times. Here we compound twice and then use a buy-and-hold investment. The resulting algorithm is denoted BAH$_W$(ANTICOR(ANTICOR)). One impact of this compounding, depicted in Figure 4, is to smooth out the anti-correlations exhibited in the stocks. It is evident that after compounding twice the returns become almost completely correlated thus diminishing the possibility that additional compounding will substantially help.[14] This idea for smoothing critical parameters may be applicable in other learning applications. The challenge is to understand the conditions and applications in which the process of compounding algorithms will have this smoothing effect.

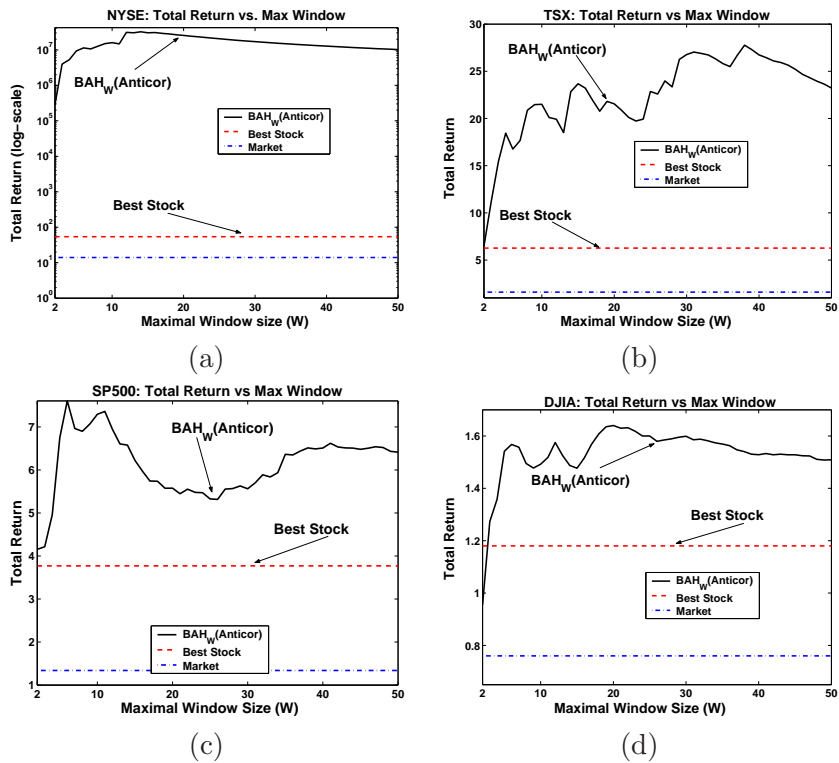

Figure 3: BAH$_W$(ANTICOR)'s total return (per \$1 investment) as a function of the maximal window $W$: NYSE (a); TSX (b); SP500 (c); DJIA (d).

## 5. An Empirical Comparison of the Algorithms

We present an experimental study of the the ANTICOR algorithm and the three online learning algorithms described in Section 3. We focus on BAH$_{30}$(ANTICOR), abbreviated by ANTI[1] and BAH$_{30}$(ANTICOR(ANTICOR)), abbreviated by ANTI[2]. Four historical datasets are used. The first NYSE dataset, is the one used in (Cover, 1991; Cover & Ordentlich, 1996; Helmbold et al., 1998) and (Blum & Kalai, 1998). This dataset contains 5651 daily prices for 36 stocks in the New York Stock Exchange (NYSE) for the twenty two year period July $3^{rd}$, 1962 to Dec $31^{st}$, 1984. The second TSX dataset consists of 88 stocks from the Toronto Stock Exchange (TSX), for the five year period Jan $4^{th}$, 1994 to Dec $31^{st}$, 1998. The third dataset consists of the 25 stocks from SP500 which (as of Apr. 2003) had the largest market capitalization. This set spans 1276 trading days for the period Jan $2^{nd}$, 1998 to Jan $31^{st}$, 2003. The fourth dataset consists of the thirty stocks composing the Dow Jones Industrial Average (DJIA) for the two year period (507 days) from Jan $14^{th}$, 2001 to Jan $14^{th}$, 2003.[15]

---

14. This smoothing effect also allows for the use of simple prediction algorithms such as "expert advice" algorithms (Cesa-Bianchi et al., 1997), which can now better predict a good window size. We have not explored this direction.

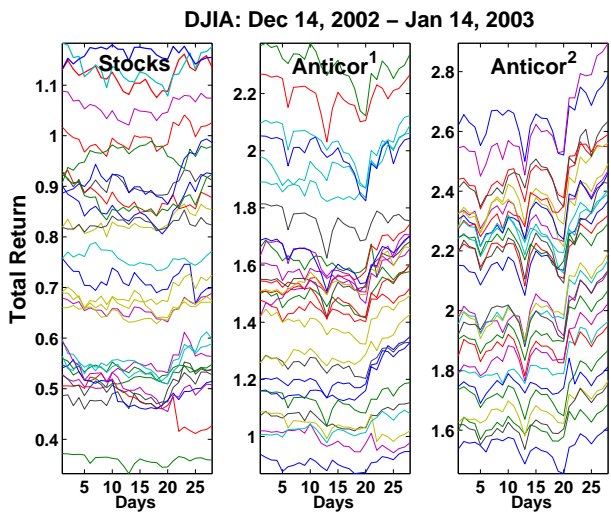

Figure 4: Cumulative returns for last month of the DJIA dataset: stocks (left panel); ANTICOR$_w$ algorithms trading the stocks (denoted ANTICOR$^1$, middle panel); ANTICOR$_w$ algorithms trading the ANTICOR algorithms (right panel).

| Algorithm | NYSE | TSX | SP500 | DJIA | NYSE$^{-1}$ | TSX$^{-1}$ | SP500$^{-1}$ | DJIA$^{-1}$ |
|---|---|---|---|---|---|---|---|---|
| MARKET (U-BAH) | 14.49 | 1.61 | 1.34 | 0.76 | 0.11 | 1.67 | 0.87 | 1.43 |
| BEST STOCK | 54.14 | 6.27 | 3.77 | 1.18 | 0.32 | **37.64** | 1.65 | 2.77 |
| CBAL* | 250.59 | 6.77 | 4.06 | 1.23 | 2.86 | **58.61** | 1.91 | 2.97 |
| U-CBAL | 27.07 | 1.59 | 1.64 | 0.81 | 0.22 | 1.18 | 1.09 | 1.53 |
| ANTI$^1$ | **17,059,811.56** | **26.77** | **5.56** | **1.59** | **246.22** | 7.12 | **6.61** | **3.67** |
| ANTI$^2$ | **238,820,058.10** | **39.07** | **5.88** | **2.28** | **1383.78** | 7.27 | **9.69** | **4.60** |
| LZ | 79.78 | 1.32 | 1.67 | 0.89 | 5.41 | 4.80 | 1.20 | 1.83 |
| EG | 27.08 | 1.59 | 1.64 | 0.81 | 0.22 | 1.19 | 1.09 | 1.53 |
| UNIVERSAL | 26.99 | 1.59 | 1.62 | 0.80 | 0.22 | 1.19 | 1.07 | 1.53 |

Table 1: Monetary returns in dollars (per \$1 investment) of various algorithms for four different datasets and their reversed versions. The winner and runner-up for each market appear in boldface. All figures are truncated to two decimals.

These four datasets are quite different in nature (the market returns for these datasets appear in the first row of Table 1). While every stock in the NYSE increased in value, 32 of the 88 stocks in the TSX lost money, 7 of the 25 stocks in the SP500 lost money and

---

15. The four datasets, including their sources and individual stock compositions can be downloaded from http://www.cs.technion.ac.il/~rani/portfolios.

25 of the 30 stocks in the "negative market" DJIA lost money. With the exception of the TSX, these data sets include only highly liquid stocks with large market capitalizations. In order to maximize the utility of these datasets and yet present rather different markets, we also ran each market in reverse. This is simply done by reversing the order and inverting the relative prices. The reverse datasets are denoted by a '-1' superscript. Some of the reverse markets are particularly challenging. For example, *all* of the NYSE$^{-1}$ stocks are going down. Note that the forward and reverse markets (i.e. U-BAH) for the TSX are both increasing but that the TSX$^{-1}$ is also a challenging market since so many stocks (56 of 88) are declining.

Table 1 reports on the total returns of the various algorithms for all eight datasets. We see that prediction algorithms such as LZ can do quite well and the more aggressive ANTI$^1$ and ANTI$^2$ have excellent and sometimes fantastic returns. Note that these active strategies beat the best stock and even CBAL* in all markets with the exception of the TSX$^{-1}$ in which case they still significantly outperform the market. The reader may well be distrustful of what appears to be such unbelievable returns for ANTI$^1$ and ANTI$^2$ especially when applied to the NYSE dataset. However, recall that the NYSE dataset consists of $n = 5651$ trading days and the $y$ such that $y^n =$ the total NYSE return is approximately 1.0029511 for ANTI$^1$ (respectively, 1.0074539 for ANTI$^2$); that is, the average daily increase is less than .3% (respectively, .75%). We observe that learning algorithms such as UNIVERSAL and EG have no substantial advantage over U-CBAL. Some previous expositions of these algorithms highlighted particular combinations of stocks where the returns significantly outperformed the best stock. But the same can be said for U-CBAL.

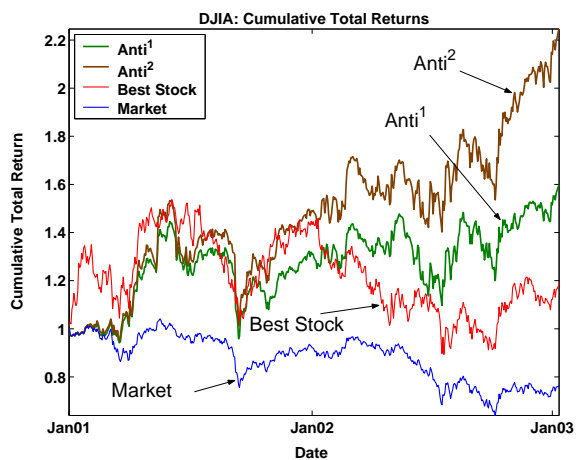

Figure 5: DJIA: Cumulative returns of of ANTI$^1$, ANTI$^2$, the best stock and a uniform BAH (the "market").

The total returns of ANTI$^1$ and ANTI$^2$ presented in Table 1 are impressive but are far from telling a complete story. Consider the graphs in figure 6. While both ANTI$^1$ and ANTI$^2$ perform well with respect to the uniform market and the best stock throughout most of the investment period, there are some periods where the cumulative return of these strategies decrease. This (not surprising) behavior indicates that there is a certain degree of risk in using these investment algorithms.

In finance the standard risk measure is the standard deviation of the return. In Table 2 we provide *annualized returns* and *risks* as well as *risk-adjusted returns* for all markets and algorithms considered here.[16] For example, the annualized return of the best stock in the DJIA set is 8.6%, its annualized risk (standard deviation) is 42% and its annualized risk-adjusted return (Sharpe ratio) is 11%.

| Algorithm | NYSE | TSX | SP500 | DJIA | NYSE$^{-1}$ | TSX$^{-1}$ | SP500$^{-1}$ | DJIA$^{-1}$ |
|---|---|---|---|---|---|---|---|---|
| MARKET | $12 \pm 14\%$ | $10 \pm 12\%$ | $5 \pm 24\%$ | $-12 \pm 24\%$ | $-9 \pm 15\%$ | $10 \pm 22\%$ | $-2 \pm 22\%$ | $19 \pm 25\%$ |
| (U-BAH) | 58% | 46% | 8% | -67% | -86% | 29% | -28% | 61% |
| BEST STOCK | $19 \pm 24\%$ | $44 \pm 55\%$ | $30 \pm 51\%$ | $8 \pm 42\%$ | $-4 \pm 21\%$ | $106 \pm 104\%$ | $10 \pm 32\%$ | $65 \pm 114\%$ |
| | 63% | 73% | 50% | 11% | -41% | 98% | 20% | 54% |
| CBAL* | $27 \pm 30\%$ | $46 \pm 40\%$ | $31 \pm 42\%$ | $11 \pm 26\%$ | $4 \pm 40\%$ | $125 \pm 78\%$ | $13 \pm 27\%$ | $71 \pm 76\%$ |
| | 78% | 106% | 65% | 27% | 1% | **156%** | 35% | 88% |
| U-CBAL | $15 \pm 13\%$ | $9 \pm 13\%$ | $10 \pm 22\%$ | $-9 \pm 25\%$ | $-6 \pm 13\%$ | $3 \pm 13\%$ | $1 \pm 21\%$ | $23 \pm 25\%$ |
| | 88% | 44% | 28% | -54% | -77% | -3% | -9% | 77% |
| ANTI$^1$ | $110 \pm 28\%$ | $93 \pm 45\%$ | $40 \pm 37\%$ | $26 \pm 35\%$ | $27 \pm 27\%$ | $48 \pm 41\%$ | $45 \pm 32\%$ | $90 \pm 31\%$ |
| | **367%** | **196%** | **95%** | **62%** | **86%** | 107% | **126%** | **277%** |
| ANTI$^2$ | $136 \pm 35\%$ | $108 \pm 60\%$ | $41 \pm 44\%$ | $50 \pm 39\%$ | $38 \pm 33\%$ | $48 \pm 46\%$ | $56 \pm 36\%$ | $113 \pm 35\%$ |
| | **370%** | **172%** | **86%** | **119%** | **101%** | 96% | **143%** | **304%** |
| LZ | $21 \pm 23\%$ | $5 \pm 25\%$ | $10 \pm 25\%$ | $-5 \pm 28\%$ | $7 \pm 21\%$ | $36 \pm 27\%$ | $3 \pm 26\%$ | $35 \pm 27\%$ |
| | 76% | 6% | 25% | -33% | 17 | **117%** | -0.8% | 112% |
| EG | $15 \pm 13\%$ | $9 \pm 13\%$ | $10 \pm 22\%$ | $-9 \pm 25\%$ | $-6 \pm 13\%$ | $3 \pm 13\%$ | $1 \pm 22\%$ | $23 \pm 25\%$ |
| | 88% | 44% | 28% | -54% | -77% | -2% | -9% | 77% |
| UNIVERSAL | $15 \pm 13\%$ | $9 \pm 13\%$ | $10 \pm 22\%$ | $-9 \pm 25\%$ | $-6 \pm 13\%$ | $3 \pm 13\%$ | $1 \pm 22\%$ | $23 \pm 25\%$ |
| | 87% | 44% | 27% | -55% | -77% | -2% | -11% | 76% |

Table 2: Annualized returns and respective annualized volatilities as well as annualized risk-adjusted returns (Sharpe Ratio) of the various algorithms over three datasets and their reversed versions. The winner and runner-up Sharpe Ratio for each market appear in boldface. All figures are truncated to two decimals.

# 6. On Commissions, Trading Friction and Other Caveats

When handling a portfolio of $m$ stocks our algorithm may perform up to $m$ transactions per day. A major concern is therefore the commissions it will incur. Within the *proportional commission* model (see e.g. Blum & Kalai, 1998; Borodin & El-Yaniv, 1998, Section 14.5.4) there exists a fraction $\gamma \in (0, 1)$ such that an investor pays at a rate of $\gamma/2$ for each buy and for each sell. Therefore, the return of a sequence $\mathbf{b}_1, \ldots, \mathbf{b}_n$ of portfolios with respect to a market sequence $\mathbf{x}_1, \ldots, \mathbf{x}_n$ is $\prod_t \left( \mathbf{b}_t \cdot \mathbf{x}_t (1 - \sum_j \frac{\gamma}{2} |\mathbf{b}_t(j) - \hat{\mathbf{b}}_t(j)|) \right)$, where

---

16. The annualized return is estimated using the geometric mean of the individual daily returns and the risk is the standard deviation of these daily returns multiplied by $\sqrt{252}$ where 252 is the assumed standard number of trading days per year. These calculations are standard. The (annualized) Sharpe ratio (Sharpe, 1975) is the ratio of annualized return minus the risk-free return (taken to be 4%) divided by the (annualized) standard deviation.

$\hat{\mathbf{b}}_t = \frac{1}{\mathbf{b}_t \cdot \mathbf{x}_t}(\mathbf{b}_t(1)\mathbf{x}_t(1), \ldots, \mathbf{b}_t(m)\mathbf{x}_t(m))$.[17] Our investment algorithm in its simplest form can tolerate very small proportional commission rates and still beat the best stock. The graphs in Figure 6 depict the total returns of BAH₃₀(ANTICOR) with proportional commission factor $\gamma = 0.1\%, 0.2\%, \ldots, 1\%$. The strategy can withstand small commission factors. For example, with $\gamma = 0.1\%$ the algorithm still beat the best stock in all four markets we consider (and it beats the market with $\gamma < 0.4\%$). Moreover it still clearly beats the market whenever $\gamma < 0.4\%$.

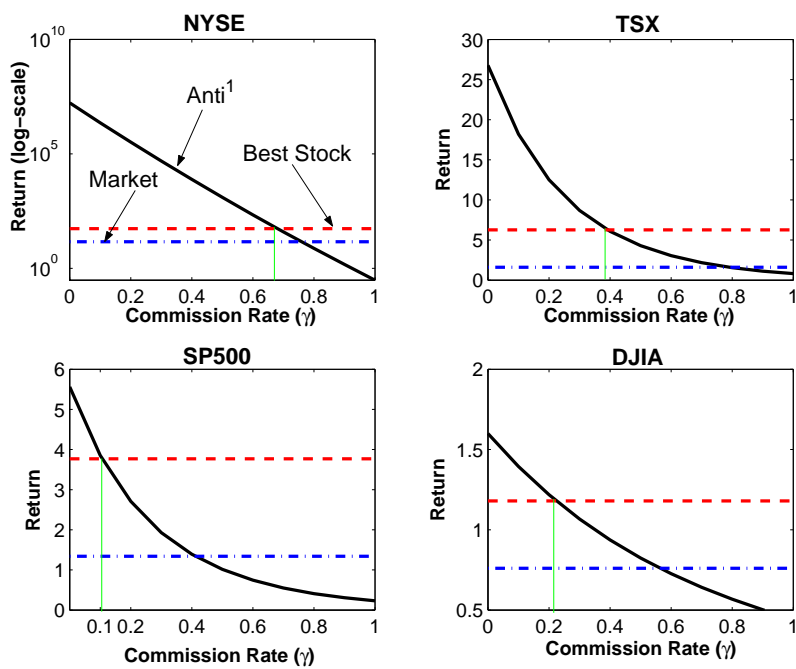

Figure 6: Total returns of BAH₃₀(ANTICOR) with proportional commissions $\gamma = 0.1\%, 0.2\%, \ldots, 1\%$.

However, some current online brokers charge very small proportional commissions, perhaps in addition to a small *flat commission rate* for all trades.[18] This means that a large investor can scale up the investment and suffer only a small proportional transaction rate.

An additional caveat is our assumption that all trades could be implemented using the closing price. While in principle there is nothing special about the closing price (i.e. our algorithms can trade at any time during the trading day) practical consideration related to dataset gathering and availability dictated the use of these prices.[19] Our algorithms assume that all portfolio adjustments are implemented using the quoted prices they receive as inputs. This means that all transactions are implemented simultaneously using the quoted prices. With current online brokers a computerized system can issue all transaction orders almost instantly but there is no guarantee that they will be all implemented instantly. This trading "friction" will necessarily generate discrepancies between the input prices and implementation prices.

---

17. We note that Blum and Kalai (1998) showed that the performance guarantee of UNIVERSAL still holds (and gracefully degrades) in the case of proportional commissions.

18. For example, on its USA site, E*TRADE (https://us.etrade.com) offers a flat fee of $10 for any trade up to 5000 shares and then $.01/share thereafter.

19. Specifically, historical closing prices are in the public domain and allow for experimental *reproducibility*. Historical intraday trading quotes can also be gathered but such data is usually protected and can be costly to obtain.

A related problem that one must face when actually trading is the difference between bid and ask prices. These bid-ask *spreads* (and the availability of stocks for both buying and selling) are functions of stock liquidity and are typically small for large market capitalization stocks. We consider here only very large market cap stocks.

Any report of abnormal returns using historical markets should be suspected of "data snooping". In particular, all of our historical data sets are conditioned on the fact that all stocks were traded every day and there were no bankrupcies or stocks that became virtually worthless in any of these data sets. Furthermore, when a dataset is excessively mined by testing many strategies there is a substantial chance that one of the strategies will be successful by simple over-fitting. Another data snooping hazard is stock selection. Our ANTICOR algorithms were fully developed using only the NYSE and TSX datasets. The DJIA and SP500 sets were obtained (from public domain sources) after the algorithms were fixed. Finally, our algorithm has one parameter (the maximal window size $W$). Our experiments clearly indicate that the algorithm's performance is robust with respect to $W$ (see, for example, Figure 4).

## 7. Concluding Remarks

Traditional work in financial economics tend to focus on the understanding of stock price determination. The main question there is: Can we predict the stock market? Judging by the extensive but inconclusive work done in financial forecasting, perhaps this is not the most beneficial question to ask. Rather, can a computer program consistently outperform the market? Besides practicality, it is clear that any successful portfolio selection algorithm is in itself a mathematical model that can provide some new intuition on stock price formation. For example, in our case, the algorithms suggest that some stock price fluctuations are sufficiently "periodic" and anti-correlated.

A number of well-respected works report on statistically robust "abnormal" returns for simple "technical analysis" heuristics, which slightly beat the market. For example, the landmark study of Brock, Lakonishok, and LeBaron (1992) apply 26 simple trading heuristics to the DJIA index from 1897 to 1986 and provide strong support for technical analysis heuristics. While *consistently* beating the market is considered a significant (if not impossible) challenge, our approach to portfolio selection indicates that beating the best stock is an achievable goal. While we have mainly focused on an idealized "frictionless setting", we believe that even in such a frictionless setting (which seems like a reasonable starting point) no such results have been previously claimed in the literature.

The results presented here raise various interesting questions. Since simple statistical relations such as correlation give rise to such outstanding returns it is plausible that various other, perhaps more sophisticated machine learning techniques, can give rise to better portfolio selection algorithms capable of larger returns and tolerating larger commissions fees.

On the theoretical side, what is missing at this point of time is an analytical model which better explains why our active trading strategies are so successful. In this regard, we are investigating various "statistical adversary" models along the lines suggested by Raghavan (1992) and Chou et al. (1995). Namely, we would like to show that an algorithm performs well (relative to some benchmark) for any market sequence that satisfies certain constraints on its empirical statistics.

One final caveat needs to be mentioned. Namely, the entire theory of portfolio selection algorithms assumes that any one portfolio selection algorithm has no impact on the market! But just like any goose laying golden eggs, widespread use will soon lead to the end of the goose. In our case, the market will quickly react to any method which does consistently and substantially beat the market.

## Acknowledgments

We thank Michael Loftus for his helpful comments. We also thank Izzy Nelken and Super Computing Inc. for their help in validating the DJIA dataset.